\documentclass[journal]{IEEEtran}
\IEEEoverridecommandlockouts                          
\usepackage{amsmath}
\usepackage{amssymb}

\usepackage{booktabs}
\usepackage{graphicx}
\usepackage{multirow}
\usepackage{listings}

\usepackage[linesnumbered,ruled,vlined]{algorithm2e}
\usepackage{hyperref}
\usepackage{todonotes}
\usepackage{cite}

\begin{document}
    \title{\LARGE \bf
Online Deep Reinforcement Learning for Autonomous UAV Navigation and Exploration of Outdoor Environments
}

\author{
    Bruna G. Maciel-Pearson$^{1}$, 
    Letizia Marchegiani$^{2}$,
    Samet Ak\c{c}ay$^{1,5}$, 
    Amir Atapour-Abarghouei$^{1,3}$,
    James Garforth$^{4}$ and 
    Toby P. Breckon$^{1}$%
\thanks{
	$^{1}$B.G. Pearson, S. Ak\c{c}ay, A.Atapour and T.P.Breckon are with Department of Computer Science, Durham University, Durham, UK (e-mail: \{
	    \href   {mailto:b.g.maciel-pearson@durham.ac.uk}
                {b.g.maciel-pearson},
		\href   {mailto:samet.akcay@durham.ac.uk}
		        {samet.akcay}, 
		\href   {mailto:amir.atapour-abarghouei@durham.ac.uk}
		        {amir.atapour-abarghouei}		        
		\href   {mailto:toby.breckon@durham.ac.uk}
		        {toby.breckon}
	\}@durham.ac.uk).
}%
\thanks{
	$^{2}$L. Marchegiani is with Department of Electronic Systems, Aalborg University, Denmark, DK (e-mail:
		\href   {mailto:lm@es.aau.dk}
		        {lm@es.aau.dk}).
}%

\thanks{
	$^{3}$ A.Atapour is with Department of Computer Science, Newcastle University, Newcastle, UK (e-mail: 
		\href   {mailto:amir.atapour-abarghouei@ncl.ac.uk}
		        {amir.atapour-abarghouei@ncl.ac.uk}
	).
}%

\thanks{
	$^{4}$ J. Garforth is with Centre for Robotics, Edinburgh University, Edinburgh, UK (e-mail:
		\href   {mailto:james.garforth@ed.ac.uk}
		        {james.garforth@ed.ac.uk}
	).
}%
\thanks{
	$^{5}$ S. Ak\c{c}ay is with COSMONiO, Durham, UK (e-mail:
		\href   {mailto:samet.akcay@cosmonio.com}
		        {samet.akcay@cosmonio.com}
	).
}%

}

\maketitle

\begin{abstract}
With the rapidly growing expansion in the use of UAVs, the ability to autonomously navigate in varying environments and weather conditions remains a highly desirable but as-of-yet unsolved challenge. In this work, we use Deep Reinforcement Learning to continuously improve the learning and understanding of a UAV agent while exploring a partially observable environment, which simulates the challenges faced in a real-life scenario. Our innovative approach uses a double state input strategy that combines the acquired knowledge from the raw image and a map containing positional information. This positional data aids the network understanding of where the UAV has been and how far it is from the target position, while the feature map from the current scene highlights cluttered areas that are to be avoided. Our approach is extensively tested using variants of Deep Q-Network adapted to cope with a double state input data. Further, we demonstrate that by altering the reward and the Q-value function, the agent is capable of consistently outperforming the adapted Deep Q-Network, Double Deep Q-Network and Deep Recurrent Q-Network. Our results demonstrate that our proposed Extended Double Deep Q-Network (EDDQN) approach is capable of navigating through multiple unseen environments and under severe weather conditions.

\end{abstract}

\thispagestyle{empty}
\pagestyle{empty}
\section{Introduction}
The use of Unmanned Aerial Vehicles (UAVs) has, in recent years, been broadly explored to aid SAR (Search and Rescue) missions in environments of difficult access for humans \cite{KARACA2018583,Silvagni2017MultipurposeUAV,Eyerman2018DJI}. In this scenario, the key advantage of a UAV is its ability to cover a larger area faster than any ground team of humans, reducing the vital search time. Although most field studies have primarily relied on the use of manually-controlled UAV, the results to date have demonstrated the importance of identifying approaches to automate search and coverage path-finding operations of a UAV specifically in SAR missions \cite{Eyerman2018DJI,MoreLifes2018DJI,adams2011survey,carlson2018adapting,Chiara2017Forestry,Sebbane2018IntelligentUAV,Kanellakis2017SurveyTrends}. 

The task of path-finding comprises a UAV following a path or a set of pre-defined waypoints until a particular destination, or a termination condition is reached \cite{maciel2018extending}. In contrast, the coverage search problem entails that the UAV will continually explore an unknown environment, regardless of the presence of a trail, while simultaneously storing a navigational route until a termination condition is reached. In both tasks, an additional objective can potentially be inserted which would allow for the detection of a possible victim, but this lies outside the scope of this work. As such, we focus on an end-to-end approach that allows the UAV to autonomously take off, find the shortest path towards the target position and land either at the target or at the nearest possible area to the target. 

 \begin{figure}
    \centering
    \includegraphics[width=\linewidth]{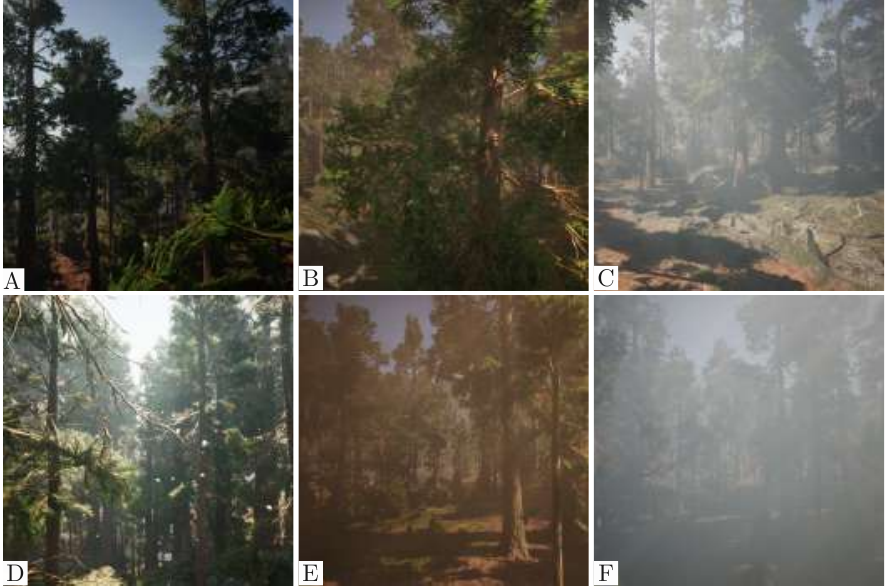}
    \caption{Field of view when flying in the forest under varying weather conditions, such as snow (light (a) and heavy (d)), dust (light (b) and heavy (e)) and fog (light (c) and heavy (f)).}
    \label{fig:weather_forest}
\end{figure}

Such an end-to-end approach would be beneficial in a search and rescue operation, as in such missions, once the victim position is estimated, it is essential to a) provide support for the victim by delivering medication, food, supplies and/or a way to communicate with the rescue team, and b) identify a navigational route that allows safe removal of the victim by the rescue team. Attempts to achieve these objectives have led to significant progress in autonomous UAV flight and path planning. However, the capability to complete an end-to-end mission that requires obstacles avoidance and deriving the shortest-path to the target whilst simultaneously mapping the environment remains an unsolved problem.

Mapping has been vastly studied over the years, resulting in a broad range of approaches that revolve around improving the global mapping of the environment either prior to or during navigation \cite{Borenstein1990, Borenstein1989}. The two most commonly used approaches for mapping are visual odometry based on optical flow and simultaneous localisation and mapping (SLAM). Variants from these approaches have demonstrated significant advancements towards navigation in changing environments \cite{Garforth2019}, \cite{Linegar2015}. However, these approaches heavily rely on camera intrinsics, which is a limiting factor when it comes to the deployment of the algorithms within UAVs with varying payload capabilities. Recently, the combination of deep learning with visual odometry \cite{Zhao2018} and SLAM have originated calibration-free algorithms. However, these approaches require a significant amount of ground truth data for training, which can be expensive to obtain and limits adaptability to other domains.

\begin{figure}
    \centering
    \includegraphics[width=\linewidth]{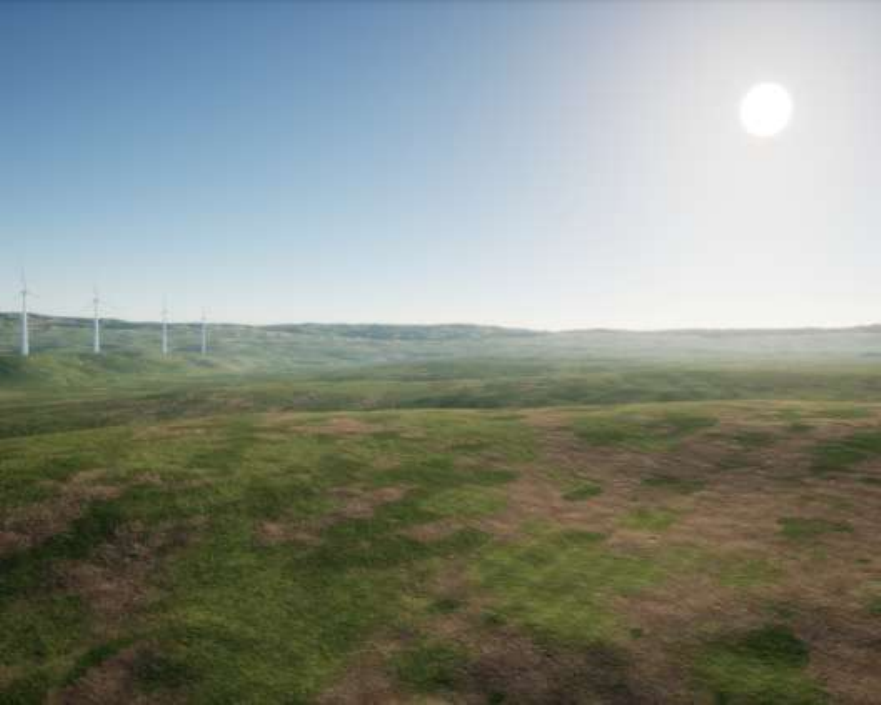}
    \caption{Illustration of the UAV field of view from the wind farm environment.}
    \label{fig:fov_farm}
\end{figure}

\begin{figure}
    \centering
    \includegraphics[width=\linewidth]{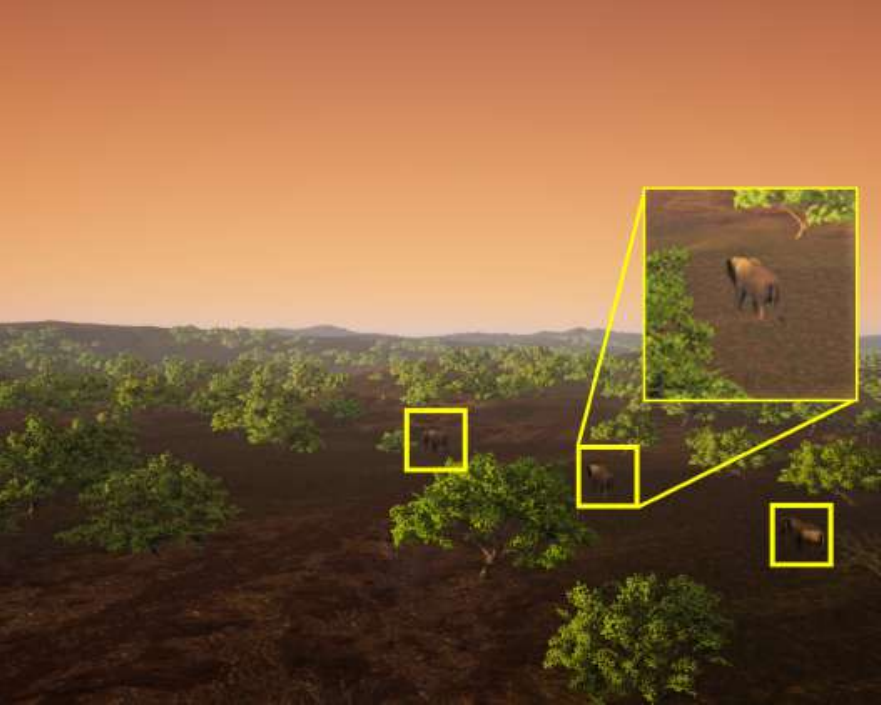}
    \caption{Illustration of the UAV field of view from the Savanna environment with moving animals.}
    \label{fig:fov_savana}
\end{figure}

In contrast, our approach can be deployed in any UAV, regardless of the camera resolution/size. Furthermore, our proposed method continuously improves its understanding of the environment over multiple flights, such that a model trained to navigate one UAV can easily be deployed in another or even multiple UAVs with different payload capabilities, without any need for offline model re-training \footnote{Source code will be made publicly available post-publication.}. This is achieved by extending Dueling Deep Q-Networks (DDQN) to fulfil the basic requirements of a multitask scenario, such as SAR missions. In particular, our approach focuses on adaptability as traversal through varying terrain formation, and vegetation is commonplace in SAR missions, and changes in terrain can significantly affect navigation capabilities. As a result, our approach is extensively tested in unseen domains and varying weather conditions such as light and heavy snow, dust and fog (Figure \ref{fig:weather_forest}) to simulate real-world exploration scenarios as closely as possible. Here, unseen domains are referred to areas previously unexplored by the UAV, which differ from the original environment on which our model, the Extended Double Deep Q-Networks (EDDQN), is trained. In this work, these domains will be areas from a dense forest environment (Figure \ref{fig:weather_forest}) that are not part of the training data; a farmland (Figure \ref{fig:fov_farm}) that presents an entirely different set of features and vegetation; and finally a Savanna environment (Figure \ref{fig:fov_savana}) with moving animals, also containing a vast amount of features that significantly differ from the training data.

Due to the nature of our tests, all the experiments are carried out in a virtual environment using the AirSim simulator \cite{shah2018airsim}. As such, our approach uses the Software in The Loop (SITL) stack, commonly used to simulate the flight control of a single or multi-agent UAV. It is also essential to observe that in a search and rescue operation, the localisation data derived from the built-in Accelerometers and Gyro combined, is usually accurate enough to allow localisation of the UAV under restricted or denied Global Positioning System (GPS) environments. As such, we use the combined readings from Accelerometers and Gyro as ground truth telemetry data, on which we base the performance evaluation for each approach. Consequently, the navigational approach presented in this paper is also non-GPS dependant.

The extensive evaluation demonstrates that our EDDQN approach outperforms contemporary state-of-the-art techniques \cite{van2016deep, hausknecht2015deep, mnih2013playing}, when tested in partially observable environments. Furthermore, the reliability and adaptability characteristics of our approach pave the way for future deployment in real-world scenarios. To best of our knowledge, this is the first approach to autonomous flight and exploration under the forest canopy that harnesses the advantages of Deep Reinforcement Learning (DRL) to continuously learn new features during the flight, allowing adaptability to unseen domains and varying weather conditions that culminate in low visibility.

    \section{Related Work}

Deep Reinforcement Learning (DRL) approaches have demonstrated significant improvements in finding the shortest path in mazes \cite{SHIN2020113064, lv2019path}, manoeuvring through urban traffic, drone racing and even military based missions \cite{yan2019towards}.

Amongst DRL techniques, Q-learning Networks have been vastly explored \cite{yan2019towards, lv2019path, krishnan2019air, hessel2018rainbow, van2016deep}, resulting in multiple variations and significant improvements. In essence, Q-learning Networks work by performing a sequence of random explorations, whereby a reward is given to each action. After a significant number of iterations, a value table is formed, and training can be done by sampling a set of actions and rewards from this value table. Conventionally, we denominate this value table as memory replay. In order to understand the direction and velocity of the agent in the environment, current  Q-learning Network variants feed the network with the past four \cite{hessel2018rainbow} or in some cases even more input images that illustrate the corresponding past actions taken by the agent. The critical drawback of this approach is the computational costs required to store and compute past actions which repeatedly appear in subsequent sub-samples during training.

Similarly, the use of Long short-term memory (LSTM) \cite{hochreiter1997long} has been vastly explored \cite{hausknecht2015deep},\cite{varior2016siamese} and although significant improvements have been achieved, our previous work \cite{maciel2019multi} demonstrates that to reach a meaningful understanding of the environment, the trained model tends to focus on high-level features from the set of input images. This subsequently results in the model becoming specialised within a single scenario/environment \cite{Zhao2017QlearningObstacleAvoidance}, thus nullifying the ability for generalisation and adaptability to changes in weather conditions or unseen environments.

In order to overcome these drawbacks, in this paper, we define the navigability problem during an SAR mission as a Partially Observable Markov Decision Process (POMDP) \cite{cassandra1998exact}, \cite{lovejoy1991survey}, in which the data acquired by the agent is only a small sample of the full searching area. Further, we propose an adaptive and random exploration approach specifically for training purpose, that is capable of self-correcting by greedily exploiting the environment when it is necessary to reach the target safely and not lose perspective of the boundaries of the search area. 

To autonomously navigate through the search area, a navigational plan needs to be devised. This navigational plan can be based either on a complete map of the environment or the perception of the local area \cite{krishnan2019air}. The former is usually applied to static environments and as such, does not require self-learning. In contrast, the latter learns environmental features in real-time, allowing for adaptability in partially or completely unknown areas \cite{Zhao2017QlearningObstacleAvoidance}. However, self-learning approaches are usually vulnerable to cul-the-sac and learning errors. These can be mitigated by adaptive mechanisms as proposed by \cite{Zhao2017QlearningObstacleAvoidance}, in which the UAV decides the best action based on the decision reached by the trap-scape and the learning modules. The former conducts a random tree search to escape the obstacles or avoid a cul-the-sac, while the latter trains the networks' model using the memory replay data. In \cite{Zhao2017QlearningObstacleAvoidance}, at each time step, the action will be derived from the computation of $Q^{N^2}$ possible actions, where $Q$ is the predicted output (\textit{Q value}) and $N^2$ represents the \textit{search area}. As such, as the size of the search area increases, so does the computational cost to define each action.

In the map-less approach of \cite{zhu2017target}, the implicit knowledge of the environment is exploited to generalise across different environments. This is achieved by implementing an actor-citric model, whereby the agent decides the next action based on its current state and the target. In contrast, our agent decides the next step by analysing the current state and a local map containing the agent's current position, navigational history and target position.

Although several well-known approaches for pose estimation are available, in this work, we refrain from incorporating pose into the pipeline and instead assume that the combined readings from the Accelerometers and Gyro offer satisfactory measurements, as demonstrated by our preliminary research \cite{maciel2019multi} and \cite{maciel2018extending}. Similarly, we do not make use of the vast work in obstacles avoidance based on depth estimation. Instead, we primarily focus on investigating how DQN variants handle continuous navigation within a partially observable environment that is continually changing. Although our training procedure is inspired by recent advances in curriculum learning \cite{krishnan2019air}\cite{fan2017learning}, in which faster convergence is achievable by progressively learning new features about the environment, we do not categorise the training data into levels of difficulties based on the loss of value function. Instead, in this work, we use stochastic gradient descent (SGD) to optimise the learning during tests using sequential and non-sequential data. However, we do progressively diversify the learning and level of difficulty by testing navigability under environmental conditions and domains unseen by the model. 

It is fair to say that our approach transfers knowledge previously extracted from one environment into another. However, we are not explicitly employing any of the current state-of-the-art transfer learning techniques available in the literature \cite{weinshall2018curriculum}\cite{yoon2019transfer}\cite{pereida2018transfer}. Our assumption is that a rich pool of features is present in the initial testing environment and our simplistic approach is capable of harnessing the best feature, similar enough to those within other environments to allow the model to converge and to adapt.

When it comes to controlling UAV actuators, the most common model-free policy gradient algorithms are deep deterministic policy gradient (DDPG) \cite{silver2014deterministic}, which have been demonstrated to be capable of greatly enhancing position control \cite{Hwangbo2017ControlQuadrotor} and tracking accuracy \cite{Wang2019DeterministicPolicy}. Trust Region Policy Optimisation (TRPO) \cite{pmlr-v37-schulman15} and Proximal Policy Optimisation (PPO) \cite{schulman2017proximal} have been shown to increase the reliability of control tasks in continuous state-action domains. However, the use of a model based on stochastic policy offers the flexibility of adaptive learning in partially or totally unknown environments, where an action can be derived from a sample of equally valid options/directions. 

Another crucial factor in safe and successful navigation is how the network perceives the environment. The computational requirements needed to support this understanding are also equally crucial, as it is important to determine whether the computation will take place on board of the UAV or by a ground control station. For instance, in \cite{qin2019autonomous}, exploration and mapping is initially performed by an unmanned ground vehicle (UGV) and complemented by an unmanned aerial vehicle (UAV). Although suitable for GPS-denied areas and fully-unknown 3D environments, the main drawback here is the fact that the UGV performs all the processing. This limits the speed and distance that can be achieved by the UAV since it needs to stay within the Wi-Fi range of the UGV. In order to overcome this drawback, significant strides have been taken towards improved onboard processing within the UAV \cite{loquercio2018dronet} but such approaches have limited generalisation capabilities. In this work, no benchmark is provided and we make use of a high-performance system, but that is primarily due to the requirements of running the simulator. In reality, our approach is designed to easily run onboard a UAV, as this is an important direction for future research within the field.
    \section{Methodology}

\begin{figure*}
    \centering
    \includegraphics[width=\linewidth]{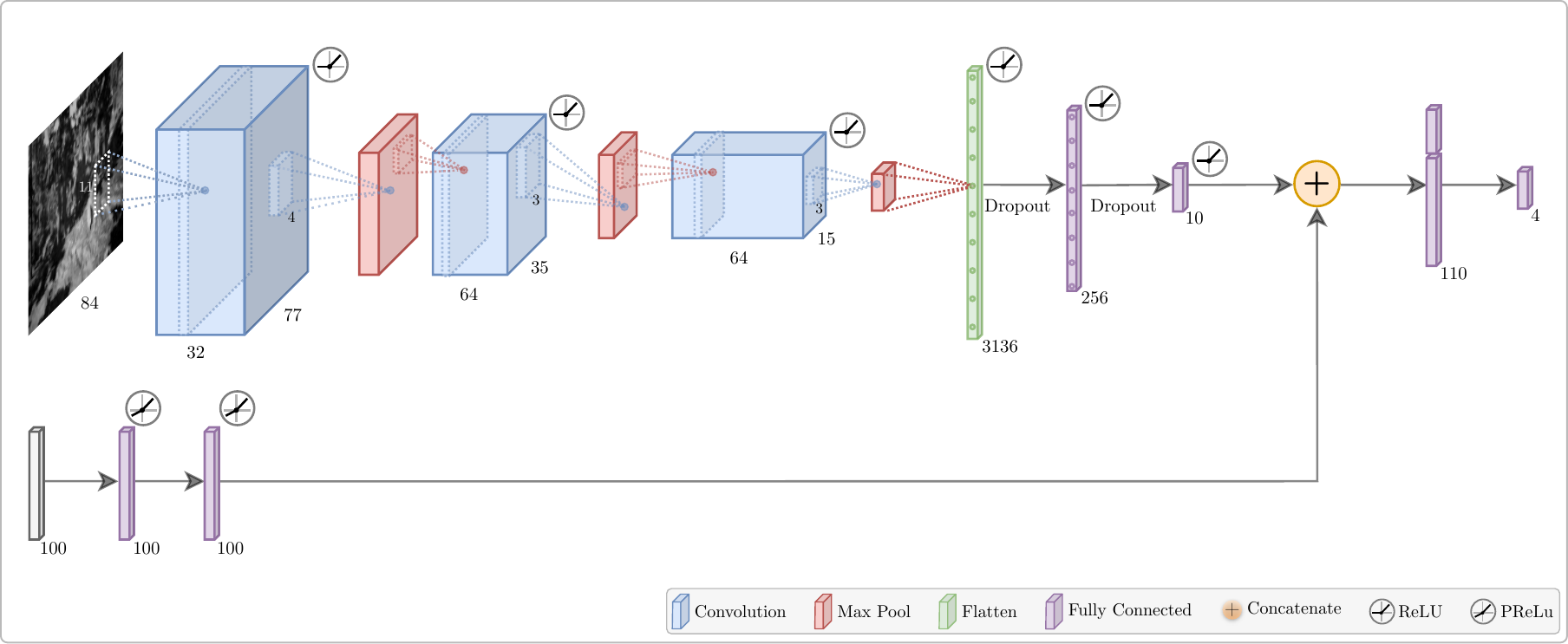}
    \caption{Model pipeline showing the double state input approach, in which the output is the estimated Q-value for each action.}
    \label{fig:pipeline}
\end{figure*}

\subsection{Mapping and Navigation}

For each mission, we define a starting position ($\hat{s}$) and a target position ($\hat{g}$), within an environment $E$, where the agent's task is to find a path $\pi$ by navigating a sequence of adjacent traversable cells such that $\pi = \pi(\hat{s},\hat{g})$. In this scenario, we assume that a path $\pi$ is feasible and that the resulting total search area is $T_{sa}= {\pi^2}/2$. Due to the nature of search and rescue missions, the position of the obstacles and possible trails within the search area are unknown. Therefore, at the start of the mission, the map, $M$, is an empty matrix of equivalent size to $T_{sa}$. Formally, we represent the environment as $E(M,\hat{s},\hat{g})$ and the objective of our model is to continuously adapt to the unknown environment during exploration, based solely on previous knowledge learned from similar environments and the need to arrive quickly and safely to the destination. As such, our model repeatedly solves $E(M,\hat{s},\hat{g})$ by recording the position of each observed obstacle in $M$ and the agent (UAV) position, as it explores $E$ to reach the target position. Further, the new understanding of $E$ is used to update the model during flight, allowing constant adaptation to new scenarios and sudden changes in weather conditions.
\begin{figure}
    \centering
    \includegraphics[width=\linewidth]{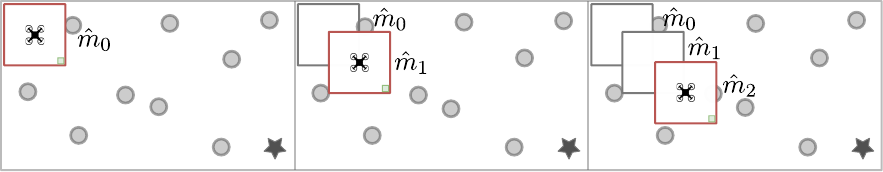}
    \caption{Abstraction of the sequence of decision maps $\hat{m}_n$ used to complete a given search mission. Grey circles represent the obstacles in the environment, while the star marks the target position ($\hat{g}$). At $\hat{m}_0$, the UAV is at its starting position ($\hat{s}$).}
    \label{fig:drone_navigation}
\end{figure}


\begin{figure}
    \centering
    \includegraphics[width=\linewidth]{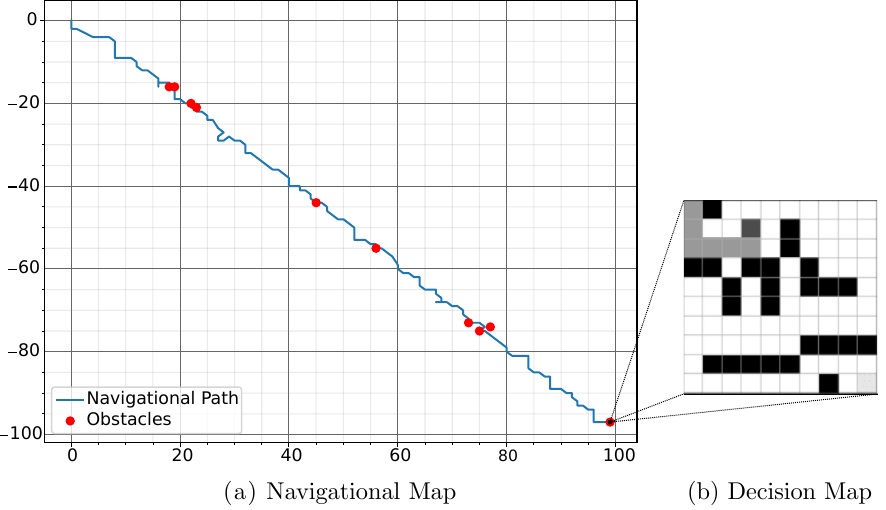}
    \caption{(a) Illustration of the navigational map, $M$, generated during mission in the predefined search area. The map is dynamically updated after completion of each decision map, $\hat{m}$. (b) Illustration of decision map $\hat{m}$, which is passed into the network during training. Black cells mark the identified obstacles, light grey represents the dominant cell which controls the heading of the UAV, medium grey denotes the visited cells and dark grey the UAV current position.}
    \label{fig:map_m}
\end{figure}

 In order to reduce the resulting computational complexity inherent to larger search areas, the proposed algorithm controls two navigational maps. The first map ($\hat{m}$), records the UAV navigation within the local environment, while the second is the representation of this navigation within the search area map, $M$ (Figure \ref{fig:map_m}). For the purpose of this work, we constrain the size of $\hat{m}$ to $10\times10$. Thus, in the first instance, the network only needs to learn how to navigate within a $100m^2$ area. The orientation of the UAV, heading towards the target position ($\hat{g}$), is controlled by the target cell $T_{cell}$. Once the UAV reaches $T_{cell}$, a new map, $\hat{m}$, is generated (Figure \ref{fig:drone_navigation}) and $M$ is updated. As a result, regardless of the size of the search area, the navigational complexity handled by the model will remain the same. To increase manoeuvrability, the agent is positioned at the centre of $\hat{m}$, allowing it to have four navigational steps in each direction. During exploration, observed obstacles are initially recorded in $\hat{m}$, in addition to the navigational route (Figure \ref{fig:map_m}). 

The mapping function specifies three levels of navigational constraints:
\begin{itemize}
    \item \textbf{Hard constraint}: the predicted navigational commands are voided, e.g. flying outside the search area or towards an obstacle.
    \item \textbf{Soft constraint}: the predicted navigational commands are permissible but punished with a lower reward, e.g. revisiting a cell.
    \item \textbf{No constraint}: the predicted navigational commands are permissible, and the mapping function records the action without any penalty, e.g. flying toward the target.
\end{itemize}

While the UAV navigation is performed in 3D (computing the $x,y,z$ axis), the grid-based representation is performed in 2D only (computing the $x,y$ axis). As such, for the purpose of representing the UAV trajectory within the grid, we define the agent position resultant from an action at time $t$ as $a_t = (x_t, y_t)$, where $x$ and $y$ are the grid cell coordinates in $M$. For simplicity, since the agent can only move one unit distance per time step, the UAV flight speed is kept at $1m/s$.

\subsection{Perception}


\begin{figure*}
    \centering
    \includegraphics[width=\linewidth]{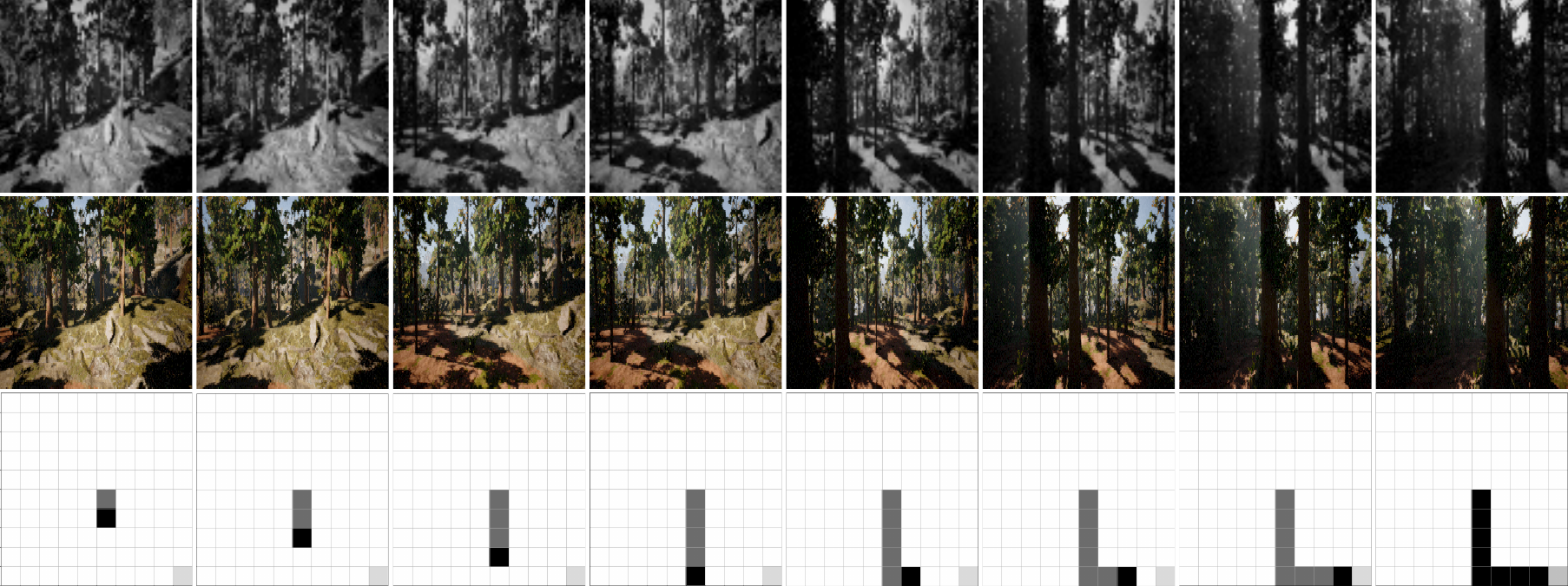}
    \caption{Sequence of double state input received by the network. The top row shows the raw image used as the input to the network, the middle row demonstrates the RGB versions of the raw images for the reader and the bottom row the decision map $\hat{m}$ at each step of the mission.}
    \label{fig:sequence_steps}
\end{figure*}

The perception of the environment is derived from the open-source AirSim simulator \cite{shah2018airsim}. Build on the Unreal Engine \cite{unrealengine}, AirSim became popular due to its physically and visually realistic scenarios for data-driven autonomous and intelligent systems. Through AirSim, a variety of environments can be simulated, including changes in luminosity and weather. Customary search missions are paused under severely low visibility due to the risks imposed on the team. Consequently, it would be of particular interest to observe the performance of our algorithm during deployment under light and heavy snow, dust and fog conditions.

It is equally essential for the approach to be able to continuously perform the mission even when drastic changes occur within the environment, as is common when the search is extended to adjacent areas, where the vegetation may significantly differ. This is especially important as it is not practical to train a model for every single environment that is potentially encountered in the  real world. With that in mind, our algorithm is trained and tested in a dense forest, but also deployed in the previously unseen environments of a farm field (Figure \ref{fig:fov_farm}) and Savannah (Figure \ref{fig:fov_savana}).

Prior to deployment, we define a search area ($SA$) and a target position ($\hat{g}$) within the AirSim environment. At each discrete time step ($t$), the agent will perform an action ($a_t$) based on what is observed in the environment; this observation is commonly defined as the state ($s_t$). However, the network is not capable of fully understanding the direction and/or velocity of navigation from one single observation. We address this issue by adding a local map, $\hat{m}$ (Figure \ref{fig:sequence_steps}), in which each cell is assigned a pixel value. As such, the position of each observed obstacle $P_o$, navigational route $P_r$, current position of the UAV $P_c$ and the remaining available cells $P_a$ are represented as a matrix of $10 \times 10$.

The input image is converted from RGB to a grey-scale image and resized to $84*84$. Both the input image and the local map are normalised to the range [-1,1]. The former allows the network to understand the advantages of avoiding cluttered areas, while the latter improves the network's understanding of the direction of flight (Figure \ref{fig:pipeline}).

A reward ($r_{t+1}$) and a discount factor $\gamma \in [0,1]$ are attributed to each triplet of state, local map and action. As a result, the transition from $s_t$ to $s_{t+1}$ can be represented by $(s_t, \hat{m}_t, a_t, r_{t+1}, \gamma,s_{t+1})$, which is subsequently added to a replay memory buffer \cite{Lin1992}.

\subsection{Reward Design}
The models' ability to learn is directly affected by the reward design. Thus, our reward design is focused on improving navigability by rewarding the agent for avoiding cluttered areas. This reduces the probability of collision, which is reinforced by blocking any cell that was previously marked as an obstacle. Once a cell is blocked, the agent cannot revisit it, and any attempt to do so is punished with a significantly lower reward value ($-1.50$). Similarly, if the agent attempts to perform an invalid action, such as flying outside the search area, the applied reward will also be low ($-0.75$). The agent is allowed to revisit previous cells within $P_r$, but this results in a decreased reward ($-0.25$). To encourage the agent to accomplish the mission faster, each hit to a valid cell results in a slight negative reward ($-0.04$). A positive reward ($1.0$) is given only when the agent reaches the target, as shown in Eqn. \ref{eq:reward}.
\begin{equation}
    r_t =
    \begin{cases}
    r^{reached} , & \text{if} \quad P_c = \hat{g} \\
    r^{blocked} ,& \text{if} \quad P_c \in P_o \\
    r^{visited} ,& \text{if} \quad p_c \in P_r \\
    r^{valid} , & \text{if} \quad p_c \in P_a \\
    r^{invalid} ,& \text{otherwise}
    \end{cases}
    \label{eq:reward}
\end{equation}

\subsection{Learning Setting}
AirSim offers three distinct modes for development: \textsc{car}, \textsc{multirotor} and \textsc{computer vision}. In both \textsc{car} and \textsc{multirotor} modes, physics inherent from the chosen environment is activated, and the vehicle behaves as it normally would in the real world. However, in the \textsc{computer vision} mode, neither the physics engines nor the vehicles are available. As results, we can restart the agent multiple times at random positions. Since the acquisition of the raw images is the same in any mode, we take advantage of the flexibility that the \textsc{computer vision} offers to train our model.

Once the model is capable of reaching the $T_{cell}$ for 50 episodes consecutively, the second testing phase is initiated. During this phase, the model is updated as the UAV exploits the environment using the \textsc{multirotor} mode. 
For both modes, the training is performed by sampling a mini-batch from the replay memory buffer and the optimisation of the network is performed by minimising the MSE.

In this work, we compare the agent's performance in executing the same mission using our adapted version of DQN \cite{mnih2013playing} and its variants DDQN \cite{van2016deep} and DRQN \cite{hausknecht2015deep}. Here, we propose and evaluate a double input state policy, in which the network receives the current state map $\hat{m}_t$ and frame $s_t$, as described in Algorithm \ref{EDDQN}. In addition, our action control policy (Section IV) reassures continuity and reliability when changes in the environment result in model instability. For simplicity, we define our adapted Q-Network as DQN* and its variants DDQN* and DRQN*. Since DQN* is based on Q-learning, its value function is updated by applying the time difference formula:

\begin{equation}
\begin{aligned}
    q_{\theta}(s_t,\hat{m}_t,a_t) = q_{\theta}(s_t,\hat{m}_t, a_t)+\alpha[ \\
    r_{t+1} + {\alpha} {max} q_{\hat{\theta}}({s_{t+1}},{\hat{m}_{t+1}},{a_{t+1}}) - q_{\theta}(s_t,\hat{m}_t,a_t)]
\end{aligned}
\end{equation}

Where $q_{\theta}(s_t,\hat{m}_t,a_t)$ represents the current state-action value function, $\alpha$ is the learning rate and ${max}q_{\hat{\theta}}({s_{t+1}},{\hat{m}_t},{a_{t+1}})$ is the estimated optimal value for transiting from $s_t$ to $s_{t+1}$. In the DQN*, the gradient of the loss is back-propagated only through the value network $\theta$, responsible for defining the action. In contrast, the target network $\hat{\theta}$, which in this case is a copy of $\theta$ is not optimised.

However, in DDQN*, the target network $\hat{\theta}$ is in fact, optimised and is responsible for evaluating the action. As such, the time difference formula in DDQN* can be defined as: 

\begin{equation}
\begin{aligned}
    q_{\theta}(s_t,\hat{m}_t,a_t) = q_{\theta}(s_t,\hat{m}_t, a_t)+\alpha[ \\
    r_{t+1} + {\alpha} q_{\hat{\theta}}({s_{t+1}},{\hat{m}_{t+1}},argmax({s_{t+1}},{\hat{m}_{t+1}},{a_{t+1}})) - \\
    q_{\theta}(s_t,\hat{m}_t,a_t)]
\end{aligned}
\label{eq:ddqn}
\end{equation}



\subsection{Network Structure}

The action value model used by the agent to predict the navigational commands for each input frame has a total of 14 layers. The first convolutional layer $(32 \times 32 \times 8)$ is followed by a max-pooling layer $(2 \times 2)$, a second convolution layer $(64 \times 64 \times 4)$, followed by a second max-pooling layer $(2 \times 2)$, a third convolutional layer $(64 \times 64 \times 3)$, followed by a third max-pooling $(2 \times 2)$, followed by a flattening operation and two dense layers. A dropout of 0.5 follows the first dense layer of size 256.  The second dense layer produces a 10-dimensional vector as the output of the network given the input image.

The local map $\hat{m}$ is also fed to the network and processed by a dense layer of size 100. The input image branch uses ReLU as the activation function in each convolutional layer. This increases the processing speed and prevents saturation. On this branch, the output will be a feature map of size 10, that will contain features indicating if the space in front of the drone is cluttered or non-cluttered. The decision map branch, on the other hand, preserves the original size through multiple convolutions and uses PReLU which, in contrast to ReLU, will allow the network to learn the negative slope of non-linearities, causing a weight bias towards zero. At the final stage, both branches are concatenated forming a layer of 110 parameters, which after been passed through a fully connected layer, it denotes the probability of each action (move left, right, forward or backwards) (Table \ref{tab:network_structure}).

\begin{table}
\caption{Network structure}
\label{tab:network_structure}
\begin{tabular}{@{}clccc@{}}
\toprule
\multicolumn{1}{l}{Branch} & Layer & \multicolumn{1}{l}{\begin{tabular}[c]{@{}l@{}}Number of\\ Neurons\end{tabular}} & \multicolumn{1}{l}{\begin{tabular}[c]{@{}l@{}}Activation\\ Function\end{tabular}} & \multicolumn{1}{l}{\begin{tabular}[c]{@{}l@{}}Kernel\\ Size\end{tabular}} \\ \midrule
I & Input Image & 84x84x1 & N/A & N/A \\
I & Convolutional & 32 & Relu & 8 \\
I & Max Pooling & N/A & N/A & 2x2 \\
I & Convolutional & 64 & Relu & 4 \\
I & Max Pooling & N/A & N/A & 2x2 \\
I & Convolutional & 64 & Relu & 3 \\
I & Max Pooling & N/A & N/A & 2x2 \\
I & Flatten & N/A & N/A & N/A \\
I & Feedforward & 256 & Relu & N/A \\
I & Feedforward & 10 & Relu & N/A \\
M & Input Map & 100 & PReLU & N/A \\
M & Feedforward & 100 & PReLU & N/A \\
IM & Concatenated{[}I,M{]} & 110 & N/A & N/A \\
Q\_values & Output & 4 & N/A & N/A \\ \bottomrule
\end{tabular}
\end{table}

The remaining DRQN* uses the same network architecture presented in Table \ref{tab:network_structure}, except for a small alteration, where after the concatenation of the outputs [I,M], an Long Short Term Memory (LSTM) layer of the same size is added. Training and tests were performed using 100 and 1000 hiding states and to retain data Time Distributed (TD) is applied to each prior layer. The full DRQN* structure can be observed in Table \ref{tab:drqn_structure}.

\begin{table}
\caption{DQRN* structure}
\label{tab:drqn_structure}
\begin{tabular}{@{}clccc@{}}
\toprule
\multicolumn{1}{l}{Branch} & Layer & \multicolumn{1}{l}{\begin{tabular}[c]{@{}l@{}}Number of\\ Neurons\end{tabular}} & \multicolumn{1}{l}{\begin{tabular}[c]{@{}l@{}}Activation\\ Function\end{tabular}} & \multicolumn{1}{l}{\begin{tabular}[c]{@{}l@{}}Kernel\\ Size\end{tabular}} \\ \midrule
I & TD: Input Image & 84x84x1 & N/A & N/A \\
I & TD: Convolutional & 32 & Relu & 8 \\
I & TD: Max Pooling & N/A & N/A & 2x2 \\
I & TD: Convolutional & 64 & Relu & 4 \\
I & TD: Max Pooling & N/A & N/A & 2x2 \\
I & TD: Convolutional & 64 & Relu & 3 \\
I & TD: Max Pooling & N/A & N/A & 2x2 \\
I & TD: Flatten & N/A & N/A & N/A \\
I & TD: Feedforward & 256 & Relu & N/A \\
I & TD: Feedforward & 10 & Relu & N/A \\
M & TD: Input Map & 100 & PReLU & N/A \\
M & TD: Feedforward & 100 & PReLU & N/A \\
IM & Concatenated{[}I,M{]} & 110 & N/A & N/A \\
IM & LSTM & 110 & Relu & N/A \\
Q\_values & Output & 4 & N/A & N/A \\ \bottomrule
\end{tabular}
\end{table}

\section{Policy Setting}
In all evaluated cases, the balance of exploration and exploitation (model prediction) obeys an $\epsilon$-greedy policy, such that:
\begin{equation}
    \pi(s_t)=
    \begin{cases}
    argmax_{a \in A} q\theta(s_t,\hat{m}_t,a), & \text{if} \quad \mu \leq \epsilon \\
    \hat{a}, & \text{otherwise}
    \end{cases}
    \label{eq:epsilon}
\end{equation}
\\
where $\mu$ is a random value generated from [0,1] per time-step, $\epsilon$ is the exploration rate and $\hat{a}$ is a random action. Although all exploration and exploitation were performed using the simulator, the proposed algorithm needs to be reliable enough for future deployment within a real UAV. As such, we added a control policy for the selection of valid actions within the $100m^2$ grid, $\hat{m}$.

\subsection{Policy for action control}
The process of continuously learning the features of the environment, to the extent that the model becomes capable of generalising to new domains is not yet optimal. We solve this issue by allowing the agent to visit a cell only if it is within $P_a$ or $P_r$. During the non-curriculum learning in the \textsc{computer vision} mode, the agent can randomly choose a valid action, when $ \mu \leq \epsilon $, otherwise it executes an action predicted by the model, without any corrections. In contrast, during curriculum learning in the \textsc{quadrotor} mode, the algorithm remains updating the model, as it learns new features. As such, to reduce the risk of collision when the model predicts an action that results in visiting a cell outside $P_a$ or $P_r$, the action is voided and the algorithm, instead, selects an action which results in visitng the adjacent cell closest to $T_{cell}$. That improves flight stability when the agent initiates flight in a new environment and/or during a change in the weather condition (Algorithm \ref{EDDQN}).

\subsection{Policy for obstacle avoidance}
Within $\hat{m}$, each cell is $1m^2$. Consequently, any object detected within a distance of less than $1m$ from the agent will be considered an obstacle. As a result, the current cell is marked as blocked, and a new direction is assigned to the agent.

\section{Extended Double Deep Q-Network}
During initial tests, we observed that multiple reward updates of action/state progressively reduce the likelihood of an action being selected given a corresponding state. As such, in contrast to EQN \eqref{eq:ddqn}, our proposed EDDQN stabilises the reward decay by altering the Q-value function such that instead of adding the reward, we subtract it, as shown in EQN \eqref{eq:eddqn}.

\begin{equation}
\begin{aligned}
    q_{\theta}(s_t,\hat{m}_t,a_t) = q_{\theta}(s_t,\hat{m}_t, a_t)+\alpha[ \\
    r_{t+1} - {\alpha} q_{\hat{\theta}}({s_{t+1}},{\hat{m}_{t+1}},argmax({s_{t+1}},{\hat{m}_{t+1}},{a_{t+1}})) - \\
    q_{\theta}(s_t,\hat{m}_t,a_t)]
\end{aligned}
\label{eq:eddqn}
\end{equation}

\begin{figure}
    \centering
    \includegraphics[width=\linewidth]{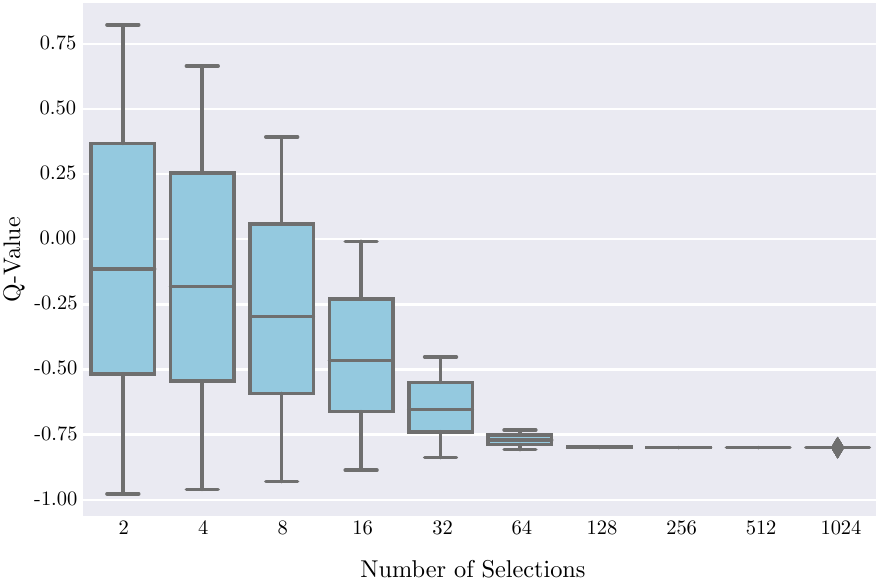}
    \caption{The box plot shows the decay of the Q-value assigned to a state over multiple updates when using the adapted DDQN*. The sample size is of 100 states, all with initial reward value equal to -0.04.}
    \label{fig:ddqn_graph}
\end{figure}

\begin{figure}
    \centering
    \includegraphics[width=\linewidth]{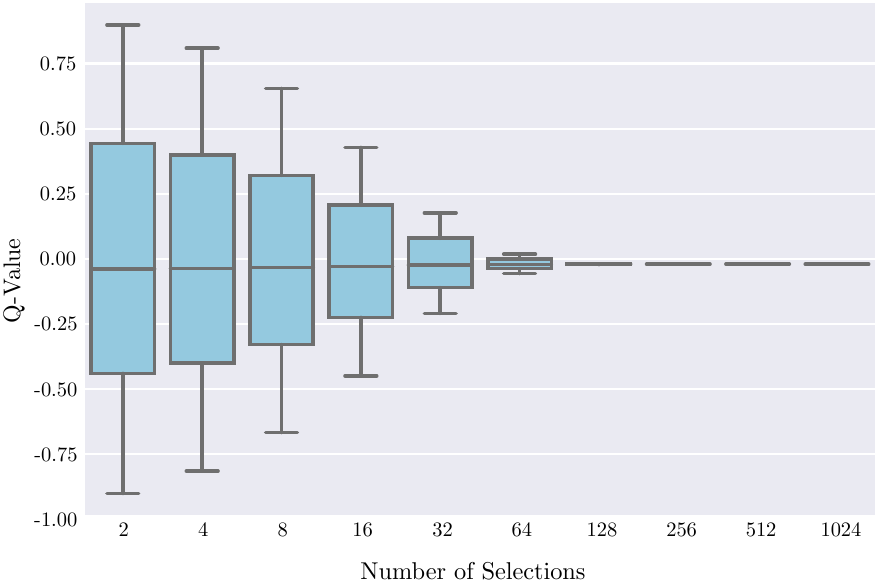}
    \caption{The box plot shows the decay of the Q-value assigned to a state over multiple updates when using our EDDQN approach. The sample size is of 100 states, all with initial reward value equal to -0.04. }
    \label{fig:eddqn_graph}
\end{figure}

The impact of this alteration can be better understood in Figures \ref{fig:ddqn_graph} and \ref{fig:eddqn_graph}, where we compare the decay of the Q-value assigned to a sample of 100 actions, all with the same initial reward value of $-0.04$, which is the value given when visiting a free cell. Furthermore, the same initial Q-value is assigned to each action during the evaluation of decay in the DDQN and EDDQN approaches.

Figure \ref{fig:ddqn_graph} demonstrates that the current Q-value function (DDQN approach) gradually degrades the value assigned to each state. As a result, if the agent visits a free cell for the first time, that set state/action will receive a high Q-value. However, as the algorithm iterates during training, every time that same set of state/action is retrieved from the memory replay a lower Q-value will be re-assigned reducing the changes of that action to be ever executed again. In the literature, this effect is mitigated because of the size of the memory replay, which is often larger than 50.000 samples, reducing significantly the changes of an action to be updated. Since we aim to further develop our EDDQN approach allowing on-board processing, we adapt the Q-value function and reduce the memory replay to only 800 samples. Combined with the desired reduction in the amount of steps per episode, it is not unusual during training for an action to be updated multiple times. As observed in Figure \ref{fig:eddqn_graph} our EDDQN approach, stabilises the Q-value assigned to each action, even over multiple updates. This, overall, improves the stability of the model by emphasising the selection of free cells over visited cells. The proposed EDDQN algorithm can be observed in detail on Algorithm \ref{EDDQN}.

\SetKwInOut{Parameter}{Initialise}
\begin{algorithm}
\label{EDDQN} 
\SetAlgoLined
\Parameter{
    Initialise memory replay buffer $B_{mr}$ to capacity $N$; Initialise the value network $\theta$ weights; Initialise the target network $\hat{\theta}$ with $\theta$ weights; Initialise $episode=0$
} 
\For{$episode < episode_{max}$}{
    get the initial observation $s_t$ \\
    \While{True}{
        get possible actions $PA$ from $(P_a,P_r)$ \\

        \eIf{$\mu < \epsilon$}
            {select a random action $a$ from $PA$\\}
            {select $a_t = argmax_{a \in A} q(s_t,\hat{m}_t,a;\theta_t)$\\}
        
        \# \textit{Apply correction to predicted action only during exploitation} \\
        
        \eIf{$exploiting==True$ and $a_t \notin PA$}
            {select action $a$ from $PA$ that is closest to $T_{cell}$}
            {}
    }
    Sample mini-batch in $B_{mr}$ \;
    Calculate the loss $(q\theta(s_i,\hat{m}_i,a_i;\theta_t)-y_i)^2$ \;
    Train and update Q networks weight $\theta_{t+1}$ \;
    Every $C$ step, copy $\theta_{t+1}$ to $\hat{\theta}$ .
}

\caption{EDDQN* Exploitation}
\end{algorithm}

    \section{Experimental Setup}
For all evaluated approaches, training is performed using adaptive moment estimation \cite{kingma2014adam} with a learning rate of $0.001$ and the default parameters following those provided in the original papers ($\beta_1 = 0.9, \beta_2 = 0.999$). As for hardware, both training and testing are performed on a Windows 10 machine, with an Nvidia GeForce RTX 2080 Ti and Intel i7-6700 CPU.

The training is performed using the AirSim \textsc{computer vision} mode \cite{shah2018airsim}, while flying in the dense forest environment. During this phase, the navigational area is restricted to $100m^2$, and the execution is completed once the agent is capable of arriving at the target position $T_{cell}$ successfully 50 consecutive times. If the agent fails to reach the target, the training will continue until a total of 1500 iterations is reached. Similar to our proposed approach, ${DRQN^*}_{100}$ and ${DRQN^*}_{1000}$ are trained by extracting sequential episodes from random positions within the memory replay. In contrast, $DQN^*$ and $DDQN^*$ learn the patterns found within the feature maps by random selection from the memory replay, which means the knowledge acquired from one frame is undoubtedly independent of any other. For all the approaches, the batch size is the same, and the agent's starting position is randomly assigned in order to increase model robustness. A detailed description of the hyper-parameters common to all approaches is presented in Table \ref{tab:hyper-parameters}.

\begin{table}[h]
\caption{Hyper-parameters used during training and testing}
\label{tab:hyper-parameters}
\resizebox{\columnwidth}{!}{%
\begin{tabular}{@{}lll@{}}
\toprule
Item & Value & Description \\ \midrule
$\epsilon$ training &  0.1     & Exploration rate. \\
$\epsilon$ testing &  0.05     & Reduced exploration rate.\\
$\gamma$ &  0.95     & Discount factor.\\
$N$ &  800     & Memory replay size.\\
$C$ &  10     & Update frequency.\\
learning rate & 0.001 & Determines how much neural net learns in each iteration. \\
batch size     &  32     &  Sample size at each step.           \\ \bottomrule
\end{tabular}%
}
\end{table}

\subsection{Evaluation Criteria}

Understanding the learning behaviour of each approach during training is critical for the deployment of a navigational system capable of self-improvement. In this context, we are primarily looking for an approach that can reach convergence quickly with as few steps as possible per episode. Fewer steps to reach the target means that the agent will not be hovering around. This is of utmost importance for autonomous flight as the UAV has a limited battery life and the key objective is to reach the target as quickly and with as little computation as possible before the battery is depleted.

Furthermore, the number of steps when correlated with the average rewards, is also an important indicator of how well the model has learned to handle the environment. High reward values demonstrate that the agent is giving preference to moving into new cells instead of revisiting cells and is therefore indicative of exploratory behaviour, which is highly desirable in search and rescue operations. As such, our evaluation is presented across two metrics: distance travelled and generalisation.

\subsubsection{Distance}
We evaluate the overall distance navigated by the agent before reaching the target and how this distance relates to the time taken. Here we are particularly interested in identifying the faster approaches.

\subsubsection{Generalisation}
The ability to generalise is categorised by observing the route chosen by the agent, time taken to reach the target and the relationship between the number of corrections and predictions. In total, ten tests are performed for each approach as detailed in Table \ref{tab:visibility}, and we aim to identify the model that completes all test. All approaches are subject to the same sequence of tests detailed in Table \ref{tab:test_sequence}.


\begin{table}
\small
\caption{Description of environmental conditions tested by each approach.}
\begin{tabular}{@{}ll@{}}
\toprule
\multicolumn{1}{c}{\textbf{Item}} & \multicolumn{1}{c}{\textbf{Description}} \\ \midrule
$F_{100}$ & Forest - searching area of $100\times100$ \\
$F_{400}$ & Forest - searching area of $400\times400$ \\
$P_{100}$ & Plain Field - searching area of $100\times100$ \\
$S_{100}$ & Savanna - searching area of $100\times100$ \\
$s_{15}$ & Light snow ($85\%$ max visibility) \\
$s_{30}$ & Heavy snow ($70\%$ max visibility) \\
$d_{15}$ & Light Dust ($85\%$ max visibility) \\
$d_{30}$ & Heavy Dust ($70\%$ max visibility) \\
$f_{15}$ & Light Fog ($85\%$ max visibility) \\
$f_{30}$ & Heavy Fog ($70\%$ max visibility) \\ \bottomrule
\end{tabular}%

\label{tab:visibility}
\end{table}

\begin{table}
\caption{Detailed sequence for testing each model}
\resizebox{\columnwidth}{!}{%
\begin{tabular}{@{}clllllllll@{}}
\toprule
\multicolumn{10}{c}{Test Sequence} \\ \midrule
I & \multicolumn{1}{c}{II} & \multicolumn{1}{c}{III} & \multicolumn{1}{c}{IV} & \multicolumn{1}{c}{V} & \multicolumn{1}{c}{VI} & \multicolumn{1}{c}{VII} & \multicolumn{1}{c}{VIII} & \multicolumn{1}{c}{IX} & \multicolumn{1}{c}{X} \\
\multicolumn{1}{l}{$F_{100}$} & $F_{400}$ & $s_{15}$ & $d_{15}$ & $f_{15}$ & $s_{30}$ & $d_{30}$ & $f_{30}$ & $P_{400}$ & $S_{400}$ \\ \bottomrule
\end{tabular}%
}

\label{tab:test_sequence}
\end{table}
    \section{Results}
In this section, we evaluate the performance of the proposed approach in predicting the next set of actions, which will result in a faster route to the target position.

\subsection{Performance during Training}

As the results of our previous work \cite{maciel2019multi} confirm, the use of deep networks capable of storing large quantities of features is only beneficial in fairly consistent environments that do not experience a large number of changes. The dense forest environment undergoes significant changes in luminosity and the position of features, since the Unreal Engine \cite{unrealengine} simulates the movement of the sun, clouds and, more importantly, the varying wind conditions which are all typical characteristics of a real living environment that is continually changing visually. These constant changes make it harder for the model to find the correlation between the feature maps and the actions executed by the agent. This results in both ${DRQN^*}_{100}$ and ${DRQN^*}_{1000}$ requiring significantly longer training time, in addition to a higher number of steps to complete a task. A detailed summary of the performance of each approach during training can be observed in Table \ref{tab:training_forest}.

\begin{table}[h]
\caption{Summary of performance for each approach during training in the Forest environment.}
\resizebox{\columnwidth}{!}{
\begin{tabular}{@{}llllll@{}}
\toprule
Approach & Steps (avg) & Completed & Reward (avg) & Time\\ \midrule
$DQN^*$ & 111.38 & 142 & -24.94 & 64.24 minutes \\ 
${DDQN^*}$ & {99.53} & 190 &  -21.80 & 1.21 hours \\ 
${DRQN^*}_{100}$ & 179.84 & 310 &  -42.44 & 10.99 hours \\ 
${DRQN^*}_{1000}$ & 180.67 & 311 &  -42.62 & 11.17 hours \\ 
\textbf{Our Approach} & \textbf{64.21} & \textbf{321} &  -42.42 & 4.11 hours\\
\bottomrule
\end{tabular}
}
\label{tab:training_forest}
\end{table}

As seen in Table \ref{tab:training_forest}, the agent trained using ${DDQN^*}$ demonstrates superior performance in this context only with relation to the overall reward (-21.80). This indicates the tendency of the agent to choose new cells over previously visited ones. While this is highly advantageous for search and rescue operations, upon more careful analysis, our approach is revealed to have a lower average step count per episode (64.21) and the highest completion rate (our approach is more likely to reach the target successfully). Another point to consider is the fact that since the starting position of the agent is randomly defined during training, the agent can be initialised near obstacles, which forces the network to revisit cells in an attempt to avoid these obstacles. This action has a significant impact on the average reward and training duration. As a result, during training, our most important evaluation factors are the average number of steps taken and the completion rate. The amount of steps taken to complete the task is indifferent to whether the cells have been previously visited or not.

Although our approach does not reach convergence as fast as $DQN$, our results (Tables \ref{tab:test_sizes},\ref{tab:test_weather} and \ref{tab:test_domains}) demonstrate that our approach is subject to the trade-off between training time and model generalisation, and a model capable of generalising to unseen environments can be highly advantageous in autonomous flight scenarios.





\subsection{Navigability within the Forest}

Although training and initial tests are performed inside the same forest environment, during testing, we extend the flight to outside the primary training area. This is done to assess whether the model is capable of coping with previously-unseen parts of the forest while diagonally traversing it. We create two separate missions and established the Euclidean distances between the take-off and the target positions for each mission. The first mission has an estimated target at $100m$ from the starting position while the second has a target at $400m$. This results in total search areas of $5,000m^2$ and $80,000 m^2$, respectively.

\begin{table}[h]
\caption{Results from autonomously exploring the forest environment with no changes in weather conditions.}
\resizebox{\columnwidth}{!}{
\begin{tabular}{@{}lllllll@{}}
\toprule
Method & Distance & Time & Obstacles & Predictions & Corrections & Random \\ \midrule
    DQN   & 100$m$ & 362.1 $sec$ & 12 & 231 & 4 & 19 \\
    DDQN   & 100$m$ & 353.0 $sec$ & 16 & 221 & 0 & 11 \\
    ${DRQN^*}_{100}$ & 100$m$ & 350.5 $sec$ & 8  & 224 & 104 & 14 \\
    ${DRQN^*}_{1000}$   & 100$m$ &28.49 $min$ & 28 & 1306 & 636 & 64\\
    Our approach & 100$m$ & 368.2 $sec$ & 8 & 233 & 117 & 21 \\\bottomrule
    DQN  & 400$m$ & 22.67 $min$ & 121 & 746 & 18 & 40 \\
    DDQN  & 400$m$ & 24.62 $min$ & 156 & 749 & 403 & 37 \\
    ${DRQN^*}_{100}$   & 400$m$ &21.99 $min$ & 119 & 786 & 288 & 40\\
    ${DRQN^*}_{1000}$   & 400$m$ &30.79 $min$ & 158 & 1082 & 229 & 56\\
    Our approach & 400$m$ & 27.00 $min$ & 142 & 954 & 476 & 41 \\
    
    \bottomrule
\end{tabular}
}
\label{tab:test_sizes}
\end{table}

This first set of tests (Table \ref{tab:test_sizes}) offer a more detailed view of the learning behaviour of each method since the starting position is the same for all tests. In this context,  ${DRQN^*}_{100}$ is the faster approach in completing both missions. However, it is essential to consider that even during the testing phase, the model is continuously being re-trained and to reduce bias toward one direction or the other, we continuously exploit the environment by randomly choosing a valid action when $\mu \leq 0.05$. As such, these randomised actions impact the overall time taken to reach the target. We can conclude, however, that the double state-input allows all approaches to navigate unexplored areas of the forest successfully, but in order to accurately identify the superior model, we need to observe the overall performance (Tables \ref{tab:test_weather}, \ref{tab:ddrn_ours} and \ref{tab:test_domains}).

\subsection{Robustness to Varying Weather}
\begin{figure*}
    \centering
    \includegraphics[width=\linewidth]{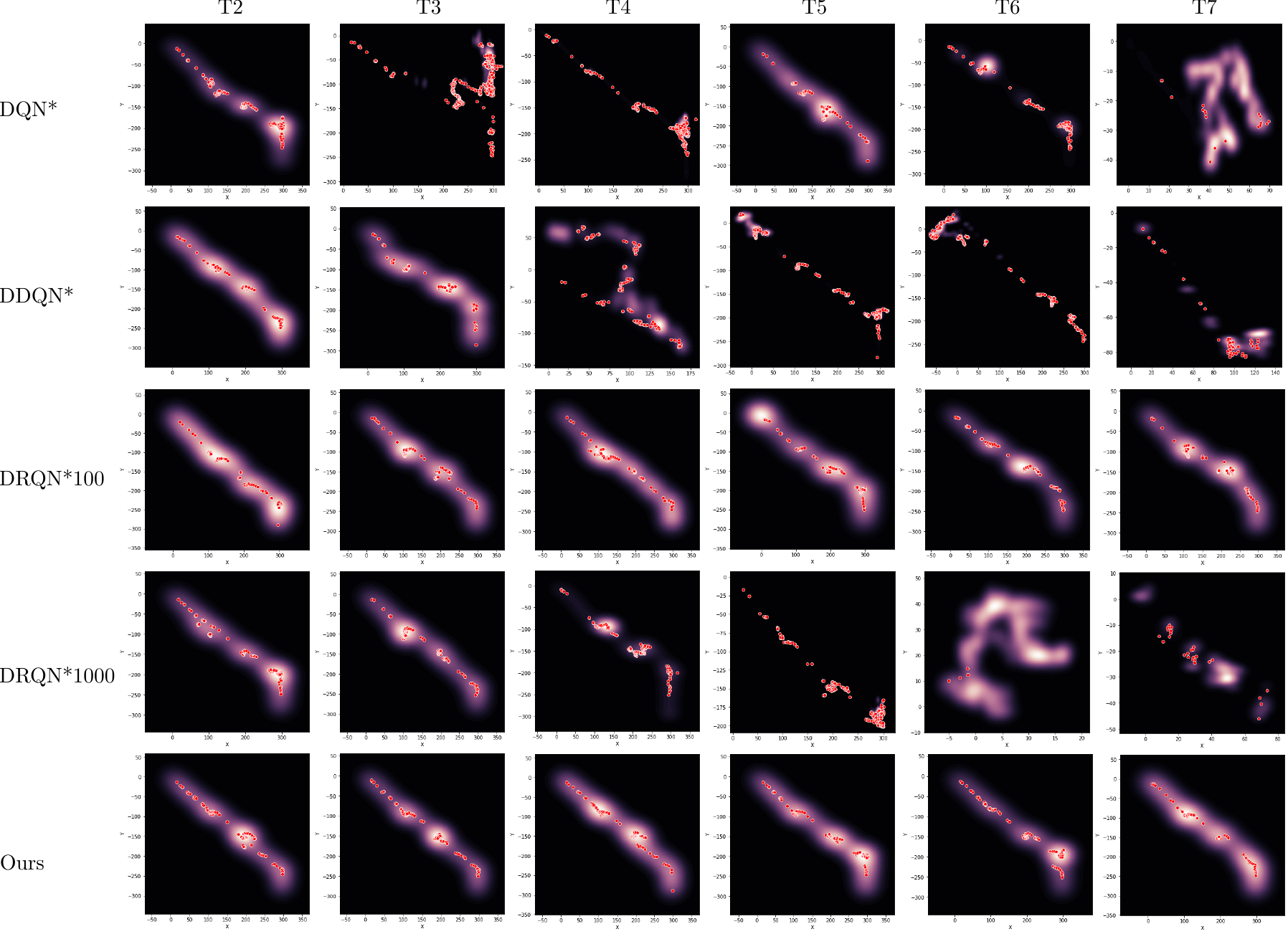}
    \vspace{-0.5cm}
    \caption{Evaluation of areas with greater flight complexity (dense white areas) and identified obstacles (red circles) position withing the 400$m$ navigational route, while flying in the default mode of $1m/s$.}
    \label{fig:weather_routes}\vspace{-0.3cm}
\end{figure*}

Here, all tests are carried out inside the dense forest environment with the target position located at a Euclidean distance of 400 meters from the take-off position. In this set of tests, we explore the behaviour of the model when flying under snowy, dusty and foggy conditions for the first time. For each condition, we compare the agent's behaviour under low and severely low visibility. The former is often very challenging, but UAV exploration of an environment with low visibility by manual control is still possible. The latter, however, is, utterly unsafe in most cases and should be avoided. The main objective of analysing the behaviour of a UAV under such constrains is to observe whether the model can give any evidence of surpassing human capabilities during a search operation. We can observe from Figure \ref{fig:weather_routes} that only ${DRQN^*}_{100}$ and our EDDQN approach are capable of navigating through the forest under all test conditions.


The noisier and the lower the visibility, the more difficult it is for the UAV to arrive at the target destination. This is observed by the significant increase in time predictions and corrections found in Table \ref{tab:test_weather} when compared with the navigational results from Table \ref{tab:test_sizes}. We further investigate the details of the performance of each approach in Table \ref{tab:test_weather} and highlight the approaches capable of completing the mission in the shortest period.

It is important to note that the increase in the number of corrections (Table \ref{tab:test_weather}) means that the agent encountered a higher number of obstacles and is thus forced to attempt to escape the boundaries of the map $\hat{m}$. This notion is further demonstrated in Figure \ref{fig:weather_routes}, where the areas of higher density (white areas) contain a greater concentration of obstacles (red circles). The only exception is when the model has the UAV continuously hovering around a small area, due to failure of the model to adapt, resulting in a higher density of flights in the presence of a small number of or no obstacles as is the case for $DQN^*$ tests T7 and T8, and ${DRQN^*}_{1000}$ test T6 (Figure \ref{fig:weather_routes}). 

\begin{table}[h]
\caption{Results from autonomously exploring the forest environment with changes in weather conditions.}
\resizebox{\columnwidth}{!}{
\begin{tabular}{@{}lllllll@{}}
\toprule
Method & Weather & Time & Obstacles & Predictions & Corrections & Random \\ \midrule
    $DQN^*$  &  snow $15\%$  & $fail$ & $fail$ & $fail$ & $fail$ & $fail$ \\
    $\mathbf{DDQN^*}$  &  snow $15\%$  & \textbf{29.23 min} & 159 & 1029 & 6 & 50 \\ 
    ${DRQN^*}_{100}$  &  snow $15\%$ & 26.38 min & 144 & 929 & 165 & 56 \\ 
    ${DRQN^*}_{1000}$  &  snow $15\%$ & 28.49 min & 152 & 990 & 112 & 60 \\ 
    Our approach & snow $15\%$ & 29.63 min & 172 & 997 & 493 & 66 \\
    \bottomrule
    $DQN^*$  &  snow $30\%$ & $fail$ & $fail$ & $fail$ & $fail$ & $fail$ \\
    $DDQN^*$  &  snow $30\%$ & $fail$ & $fail$ & $fail$ & $fail$ & $fail$\\ 
    $\mathbf{{DRQN^*}_{100}}$  &  snow $30\%$ & \textbf{28.21 min}   & 124 &  1077  & 266 & 51 \\ 
    ${DRQN^*}_{1000}$  &  snow $30\%$ & $fail$ & $fail$ & $fail$ & $fail$ & $fail$ \\
    Our approach & snow $30\%$ & 30.96 min & 173 & 1058 & 525 & 61 \\
    \bottomrule
    $DQN^*$  &  dust $15\%$ & $fail$ & $fail$ & $fail$ & $fail$ & $fail$ \\
    $DDQN^*$  &  dust $15\%$ & $fail$ & $fail$ & $fail$ & $fail$ & $fail$\\ 
    ${DRQN^*}_{100}$  &  dust $15\%$ &  29.48 min & 151 & 1082 & 364 & 59 \\ 
    ${DRQN^*}_{1000}$  &  dust $15\%$ & 45.65 min & 195 & 1735 & 529 & 106 \\ 
    \textbf{Our approach} & dust $15\%$ & \textbf{26.44 min} & 137 & 931 & 458 & 49 \\
    \bottomrule
    $DQN^*$  &  dust $30\%$ & $fail$ & $fail$ & $fail$ & $fail$ & $fail$ \\
    $\mathbf{DDQN^*}$  &  dust $30\%$ & \textbf{23.67 min} & 95 & 894 & 14 & 56 \\ 
    ${DRQN^*}_{100}$  &  dust $30\%$ & 29.55 min & 177 & 1029 & 329 & 56 \\ 
    ${DRQN^*}_{1000}$  &  dust $30\%$ & $fail$ & $fail$ & $fail$ & $fail$ & $fail$ \\ 
    Our approach & dust $30\%$ & 23.85 min & 114 & 862 & 427 & 35 \\
    \bottomrule
    $\mathbf{DQN^*}$  &  fog $15\%$ & \textbf{28.11 min} & 414 & 990 & 3 & 58 \\
    $DDQN^*$  &  fog $15\%$ & $fail$ & $fail$ & $fail$ & $fail$ & $fail$ \\ 
    ${DRQN^*}_{100}$  &  fog $15\%$ & 32.73 min & 151 & 1213 & 511 & 61 \\ 
    ${DRQN^*}_{1000}$  &  fog $15\%$ &  $fail$ & $fail$  & $fail$ & $fail$ & $fail$  \\ 
    Our approach & fog $15\%$ & 30.46 min & 114 &1054 & 523 & 59 \\
    \bottomrule
    $DQN^*$  &  fog $30\%$ & $fail$ & $fail$ & $fail$ & $fail$ & $fail$ \\
    $DDQN^*$  & fog $30\%$ & 35.33 min & 238 & 1186 & 63 & 53 \\ 
    ${DRQN^*}_{100}$  & fog $30\%$ &  35.64 min & 206 & 1262 & 443 & 74 \\ 
    ${DRQN^*}_{1000}$  &  fog $30\%$ & $fail$ & $fail$ & $fail$ & $fail$ & $fail$ \\ 
   \textbf{Our approach} & fog $30\%$ & \textbf{29.95 min} & 166 & 1052 & 517 & 51 \\
    \bottomrule
    
\end{tabular}
}
\label{tab:test_weather}
\end{table}

Since only ${DRQN^*}_{100}$ and our approach are successful in completing all tests under varying weather conditions (Table \ref{tab:test_weather}), we further evaluate their understanding of the environment (Table \ref{tab:ddrn_ours}). Here, the capability of a model to adapt to the environment is indicated by a lower average step count and a higher reward rate.

\begin{table}[]
\small
\caption{Overall performance under varying weather conditions.}
\resizebox{\columnwidth}{!}{
\label{tab:ddrn_ours}
\begin{tabular}{@{}cllll@{}}
\toprule
Test & Approach & Steps (avg) & Reward (avg) & Time \\ \midrule
\multirow{2}{*}{III} & $\mathbf{{DRQN^*}_{100}}$ & \textbf{7.35} & \textbf{0.6315}  & \textbf{26.38 minutes} \\
 & Our Approach & 7.5 & 0.4162   & 29.63 minutes \\ \midrule
\multirow{2}{*}{IV} & ${DRQN^*}_{100}$  & 8.2 & 0.4452  & 29.48 minutes \\
 &  \textbf{Our Approach}  & \textbf{7.45} & \textbf{0.4883}  & \textbf{26.44 minutes} \\ \midrule
\multirow{2}{*}{V} & ${DRQN^*}_{100}$ & 8.8 & 0.2573  & 32.73 minutes \\ 
 & \textbf{Our Approach} & \textbf{7.5} & \textbf{0.5079}  & \textbf{30.46 minutes} \\ \midrule
\multirow{2}{*}{VI} & ${DRQN^*}_{100}$ & 9.25 & 0.1642  & \textbf{28.21 minutes} \\
 & \textbf{Our Approach} & \textbf{7.48} & \textbf{0.4617}  & 30.96 minutes \\ \midrule
\multirow{2}{*}{VII} & ${DRQN^*}_{100}$ & 7.41 & 0.6467  & 29.55 minutes \\
 & \textbf{Our Approach} & \textbf{7.23} & \textbf{0.5224}  & \textbf{23.85 minutes} \\ \midrule
\multirow{2}{*}{VIII} & ${DRQN^*}_{100}$ & 8.03 & 0.4875  & 35.64 minutes \\
 & \textbf{Our Approach} & \textbf{7.36} & \textbf{0.5365}  & \textbf{29.95 minutes} \\  
 \bottomrule
\end{tabular}
}
\end{table}

Table \ref{tab:ddrn_ours} demonstrates that, in general, our approach outperforms ${DRQN^*}_{100}$ by requiring significantly fewer steps per episode to reach the target, $T_{cell}$, and with overall higher reward per episode, which means our approach gives preference to exploration rather than revisiting previous areas.

\subsection{Generalisation to Unseen Domains}
Our final set of tests explore the capability of the model to handle different environments (Table \ref{tab:test_domains}). As such, tests are also carried in a wind farm (Figure \ref{fig:fov_farm}) and a Savanna (Figure \ref{fig:fov_savana}) environment. While the former is expected to be easily navigable, the latter adds an extra level of difficulty since it contains moving animals that frequently appear in the agent's field of view (Figure \ref{fig:fov_savana}).

\begin{table}[]
\caption{Results from autonomously exploring varying domains with no changes in weather conditions.}
\resizebox{\columnwidth}{!}{
\begin{tabular}{@{}lllllll@{}}
\toprule
Method & Domain & Time & Obstacles & Predictions & Corrections & Random \\ \midrule
    $DQN^*$ & Plain & 23.29 min & 4 & 1115 & 34 & 55 \\
    $DDQN^*$ & Plain & 35.56 min & 4 & 1740 & 305 & 84 \\
    $\mathbf{{DRQN^*}_{100}}$ & \textbf{Plain} & \textbf{12.87 min} & \textbf{3} & \textbf{577} & \textbf{256} & \textbf{24} \\
    ${DRQN^*}_{1000}$ & Plain & 12.90 min & 0 & 583 & 208 & 29 \\
    Our approach & Plain &  13.97 min & 0 & 612 & 307 & 21   \\\bottomrule
    $\mathbf{DQN^*}$ & \textbf{Savana} & \textbf{12.67 min} & \textbf{0} & \textbf{575} & \textbf{57} & \textbf{31} \\
    $DDQN^*$ & Savana & 12.97 min & 0 & 589 & 5 & 27 \\
    ${DRQN^*}_{100}$ & Savana & 12.90 min & 0 & 583 & 208 & 29 \\
    ${DRQN^*}_{1000}$ & Savana & $fail$ & $fail$ & $fail$ & $fail$ & $fail$ \\
     Our approach & Savana & 13.34 min & 0 & 583 & 292 & 30 \\ \bottomrule
\end{tabular}
}
\label{tab:test_domains}
\end{table}

Our EDDQN approach successfully generalises to both environments. No drop in performance is observed by having moving animals within the scene.

Overall, adapting each network to receive a double state-input significantly reduces the expected computation. Since instead of feeding the network with 28,224 features (84x84x4) as proposed by \cite{mnih2013playing}, we input only 7,156 features (84x84+100). Further, the use of a local map offers a reliable mechanism to control the drone position within the environment and reduces the complexity inherent in handling broader areas. Our results indicate that the proposed Feedforward EDDQN approach offers a reliable and lighter alternative to the Recurrent Neural Network-based approaches of ${DRQN^*}_{100}$ and ${DRQN^*}_{1000}$, as it is more suitable for on-board processing.

    \section{Conclusion}

In Search and Rescue missions, speed is often crucial for the success of the mission, but is not the only factor. Severe weather conditions that may result in low visibly and often the varying vegetation/terrain that needs to be traversed also have significant impacts on the outcome of the mission. These, combined, significantly increase the difficulty of any mission. In this work, we demonstrate that by using deep reinforcement learning with a double input state, it is possible to exploit the capability for obstacle avoidance derived from the raw image in addition to the navigational history that is registered on the local map.

Our results demonstrate that our approach is capable of completing test missions within the current battery time available in most commercial drones, which is, of course, of significant importance in any search and rescue operation. Furthermore, we present an EDDQN technique that consistently overcomes and adapts to varying weather conditions and domains, paving the way for future development and deployment in real-world scenarios. Since our approach significantly reduces the amount of data processed during each mission, our next step is to continuously improve navigability by reducing the number of corrections and steps required to complete each search mission. In addition, we aim to adapt this work to onboard processing on a real drone with increased flight speed.

\addtolength{\textheight}{-12cm}   








    \bibliographystyle{IEEEtran}
\bibliography{bibliography.bib}


\end{document}